\documentclass[conference]{IEEEtran}
\IEEEoverridecommandlockouts
\usepackage{cite}
\usepackage{amsmath,amssymb,amsfonts}
\usepackage{algorithmic}
\usepackage{graphicx}
\usepackage{textcomp}
 \usepackage{float} 
 \usepackage{tabularx}
 \usepackage{booktabs} 
 \usepackage{placeins}
 \usepackage{changepage}
\usepackage{subcaption}
\usepackage{url}

\usepackage{xcolor}
\def\BibTeX{{\rm B\kern-.05em{\sc i\kern-.025em b}\kern-.08em
    T\kern-.1667em\lower.7ex\hbox{E}\kern-.125emX}}
\begin{document}

\title{A New Dataset and Performance Benchmark for Real-time Spacecraft Segmentation in Onboard Computers}

\author{\IEEEauthorblockN{Jeffrey Joan Sam\IEEEauthorrefmark{1},
Janhavi Sathe\IEEEauthorrefmark{1},
Nikhil Chigali\IEEEauthorrefmark{1}, 
Naman Gupta\IEEEauthorrefmark{1},
Radhey Ruparel\IEEEauthorrefmark{1}, 
Yicheng Jiang\IEEEauthorrefmark{1}, \\
Janmajay Singh\IEEEauthorrefmark{1}, 
James W. Berck\IEEEauthorrefmark{2}, and
Arko Barman\IEEEauthorrefmark{3}}
\IEEEauthorblockA{\IEEEauthorrefmark{1}Department of Computer Science, Rice University, Houston TX, USA}
\IEEEauthorblockA{\IEEEauthorrefmark{2}Johnson Space Center, NASA, Houston TX, USA}
\IEEEauthorblockA{\IEEEauthorrefmark{3}Data to Knowledge Lab, Rice University, Houston TX, USA \\
Email: arko.barman@rice.edu}}


\maketitle

\begin{abstract}
Spacecraft deployed in outer space are routinely subjected to various forms of damage due to exposure to hazardous environments. In addition, there are significant risks to the subsequent process of in-space repairs through human extravehicular activity or robotic manipulation, incurring substantial operational costs. Recent developments in image segmentation could enable the development of reliable and cost-effective autonomous inspection systems. While these models often require large amounts of training data to achieve satisfactory results, publicly available annotated spacecraft segmentation data are very scarce. Here, we present a new dataset of nearly 64k annotated spacecraft images that was created using real spacecraft models, superimposed on a mixture of real and synthetic backgrounds generated using NASA's TTALOS pipeline. To mimic camera distortions and noise in real-world image acquisition, we also added different types of noise and distortion to the images. Our dataset includes images with several real-world challenges, including noise, camera distortions, glare, varying lighting conditions, varying field of view, partial spacecraft visibility, brightly-lit city backgrounds, densely patterned and confounding backgrounds, aurora borealis, and a wide variety of spacecraft geometries. Finally, we finetuned YOLOv8 and YOLOv11 models for spacecraft segmentation to generate performance benchmarks for the dataset under well-defined hardware and inference time constraints to mimic real-world image segmentation challenges for real-time onboard applications in space on NASA's inspector spacecraft. The resulting models, when tested under these constraints, achieved a Dice score of 0.92, Hausdorff distance of 0.69, and an inference time of about 0.5 second. The dataset and models for performance benchmark are available at \url{https://github.com/RiceD2KLab/SWiM}.

\end{abstract}


\section{Introduction}
The rapid expansion of human orbital infrastructure has created an urgent unmet need for autonomous spacecraft inspection systems. As satellite constellations grow and deep space missions proliferate, inspector spacecraft, specifically designed to inspect other spacecraft in real-time, must reliably analyze diverse targets ranging from defunct satellites to interplanetary probes. This operational reality demands robust computer vision systems capable of adapting to unknown spacecraft architectures under the extreme constraints of space-grade computing hardware -- a challenge compounded by the variations in lighting, motion artifacts, and sensor noise endemic to orbital imaging.

In general, the hardware and computational constraints for real-time spacecraft segmentation on a flight computer on board an inspector spacecraft can be summarized as:
\begin{itemize}
    \item Hardware constraint: Inference on 4-core CPU with less than 4GB RAM.
    \item Inference time constraint: Inference time, $\tau_{inf} < 0.95$ second.
\end{itemize}

Current spacecraft segmentation methods face three interrelated limitations. First, existing datasets such as NASA's PoseBowl dataset~\cite{drivendata2024pose} and the annotated Spacecrafts dataset~\cite{hoang2021spacecraftdataset} lack diversity in spacecraft geometries and background conditions. Second, no established benchmarks evaluate performance under hardware constraints equivalent to flight computers (less than 4GB RAM, CPU-only inference) and inference time constraints (inference time less than $0.95$ second). Third, conventional metrics such as the Dice coefficient fail to capture boundary localization precision critical for proximity operations. These gaps hinder the development of algorithms that can be deployed on resource-constrained orbital platforms.

To address these challenges, we introduce a dual-methodology dataset synthesis approach. Building on the PoseBowl and Spacecrafts datasets, we created our dataset, Spacecraft With Masks (SwiM), through two complementary strategies: (1) superimposing existing spacecraft images on augmented open-source backgrounds with photometric/geometric distortions to mimic real-world noise and distortions in image acquisition in space, and (2) generating synthetic samples via NASA's TTALOS (Toolset for Training and Labeling in an Optical Simulator) pipeline, which integrates astrophysical backgrounds generated using stable diffusion with procedurally rendered 3D spacecraft models. This hybrid methodology achieves unprecedented diversity in spacecraft poses, lighting conditions, and environmental contexts while maintaining physical precision and simulating camera distortions and noise. To the best of our knowledge, our SWiM dataset, consisting of nearly 64k images with annotations, is the largest and most comprehensive spacecraft segmentation dataset to date.

While large datasets, such as SPEED+~\cite{speed_plus}, are available for pose estimation of spacecraft, these datasets do not include the pixel-level segmentation masks of spacecraft for training models. Furthermore, these datasets have limited diversity of spacecraft images. For example, SPEED+ consists of images of only a single spacecraft (Tango) generated using an OpenGL-based emulator (SLAB), whereas our dataset includes multiple spacecraft generated via the TTALOS pipeline.

The computational constraints of orbital hardware necessitate architectures balancing inference time and precision. We adapt the You Only Look Once (YOLO) framework\cite{redmon2016you}, renowned for its real-time inference capabilities, for spacecraft segmentation tasks. While other CNN-based object detection and segmentation models, such as Mask-RCNN~\cite{maskrcnn} excel in similar tasks, their computational requirements and inference time often exceed the capabilities of resource-constrained flight-grade hardware. The unified detection-segmentation approach in recent variants of YOLO, such as YOLOv8~\cite{ultralytics_yolov8} and YOLOv11~\cite{ultralytics_yolo11}, enables efficient processing on our target onboard flight computer with 4GB RAM, which is used by NASA for on-board flight computers in inspector spacecraft.

Our work presents four key contributions to real-time autonomous spacecraft inspection in onboard flight computers:
\begin{itemize}
    \item By creating the first large benchmark dataset explicitly designed for hardware-constrained segmentation, we enable standardized evaluation of orbital vision systems.
    \item Our problem formulation codifies real-world deployment requirements through quantifiable constraints on hardware (inference on CPU with $<$4 GB RAM only) and inference latency ($<$0.95s).
    \item We introduce a dual-metric evaluation protocol combining region-based (Dice coefficient) and boundary-aware (Hausdorff distance) metrics, addressing the limitations of single-criterion assessments in prior work.
    \item Finally, we report performance benchmarks for our dataset under the defined constraints for future research.
\end{itemize}
These advances provide a framework for developing vision-based systems that meet the stringent demands of in-orbit autonomy.

\section{Literature Review}
 
\subsection{Datasets}

There are only a few datasets for spacecraft detection and segmentation. We note that none of these datasets provide a single segmentation mask for the entire spacecraft, to the best of our knowledge.

\textit{PoseBowl Detection Dataset} (henceforth referred to as \textit{PoseBowl})~\cite{drivendata2024pose} is part of the NASA PoseBowl challenge and contains images and bounding box data to train object detection models for generic spacecraft in images. The dataset simulates images taken from the perspective of a small inspector spacecraft using 2D projections of 3D spacecraft models in Blender, a free and open-source 3D creation software~\cite{blender}. Camera distortions such as motion blur, hot pixels, and random noise were introduced to simulate real-world image acquisition.

\textit{Spacecrafts}~\cite{hoang2021spacecraftdataset} includes 3,116 real and synthetic images of spacecraft. It is designed for both object detection and semantic segmentation tasks, thus providing annotations in the form of both bounding boxes and separate segmentation masks for the spacecraft's body, solar panels, and antennae.

\begin{table}[t]
\renewcommand{\arraystretch}{1.3} 
\caption{Comparison of Spacecrafts and PoseBowl Detection Datasets}
\centering
\begin{tabular}{|p{1.5cm}|p{3cm}|p{3cm}|} 
\hline
\textbf{Dataset}  & \textbf{PoseBowl} & \textbf{Spacecrafts} \\ \hline
\textbf{\# Images} & 25,801 images & 3,116 images  \\ \hline
\textbf{Resolution} & 1024$\times$1280 pixels & 1280$\times$720 pixels \\ \hline
\textbf{Data Format} & RGB images with bounding box annotations & RGB images with segmentation masks and bounding box annotations \\ \hline
\textbf{Segmentation Masks}  & N/A & Three parts segmented (body, solar panel, antenna) \\ \hline
\textbf{Train/Test Splits}  & Public and private datasets for training and evaluation & 403 fine masks for training, 2114 coarse masks for training, and 600 for validation\\ \hline
\textbf{Memory Size}  & 1.82 GB (smallest: 8 KB, largest: 250 KB) & 2.16 GB (smallest: 49 KB, largest: 2.4 MB)\\ \hline
\end{tabular}
\label{table:dataset_comparison}
\end{table}

Both datasets were curated to simulate real-world scenarios encountered by inspector spacecraft deployed for close-up observations of other orbiting spacecraft. Table \ref{table:dataset_comparison}  presents a comparison of the datasets, including information on the number of images, memory size, resolution, and additional details.

Recently, other datasets have been generated and made available~\cite{10115564, 10380525}. However, these datasets are completely synthetic or only have a few spacecraft models in the images.

\subsection{Models}

\begin{figure}[ht]  
    \centering
    \includegraphics[width=\columnwidth]{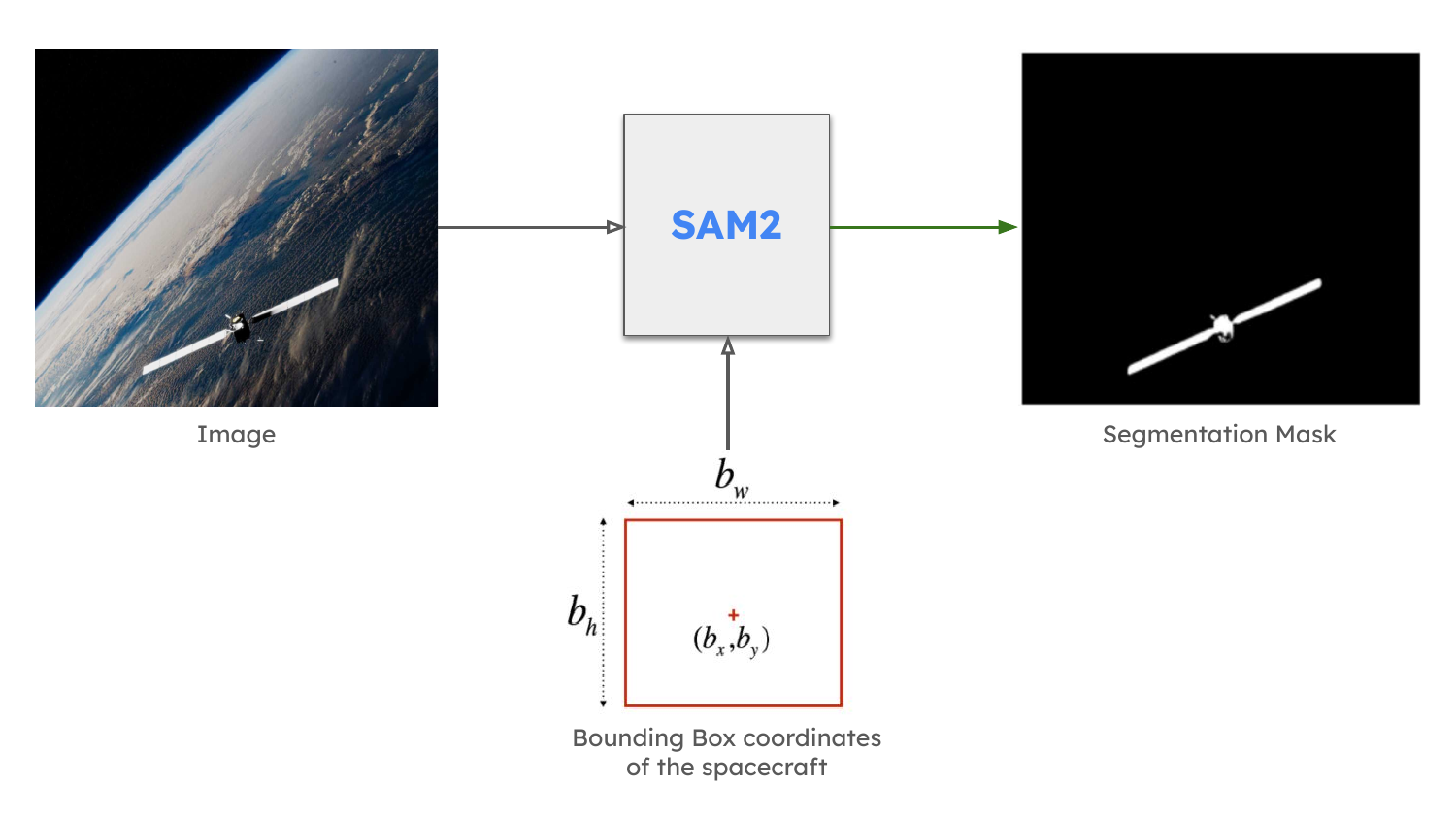}  
    \caption{Example showing segmentation mask generation using SAM 2 for an image in the \textit{PoseBowl} dataset.}
    \label{fig:posebowl_example}
\end{figure}

Due to the lack of available spacecraft datasets with annotated segmentation masks, very few benchmark performance results or novel segmentation methods have been developed for spacecraft segmentation. Performance benchmarks on these datasets include the use of different models, such as DeepLabV3+ Xception~\cite{DBLP:journals/corr/abs-1802-02611}, ResNeSt~\cite{9857221}, HRNet~\cite{9052469}, and a novel method, SpaceSeg~\cite{liu2025spaceseghighprecisionintelligentperception}, an encoder-decoder architecture with attention refinement. Other similar problems in space travel and in-flight spacecraft inspection, such as spacecraft pose estimation~\cite{10682976}, also leverage spacecraft segmentation as a whole or segmentation of certain components of spacecraft.

We note here that, to the best of our knowledge, none of the available segmentation performance benchmarks has been reported to satisfy strict hardware or inference time constraints, which are critical for the deployment of a segmentation model on board a spacecraft for real-time in-flight segmentation.

\section{Spacecraft With Masks (SWiM) Dataset}

We introduce our spacecraft segmentation dataset, SwiM (Spacecraft With Masks), which includes preprocessed images and newly-generated annotations from the two aforementioned public datasets, \textit{PoseBowl} and \textit{Spacecrafts}, as well as synthetically generated images using two different methods. The SWiM dataset is available in two versions -- baseline and augmented. Further, the two versions of the dataset are pre-split into training, validation, and test partitions for ease of benchmarking and model performance comparisons.

First, we discuss the preprocessing steps and mask generation for the pre-existing datasets. Then, we discuss the synthetic image generation methods. Finally, we describe the composition of the two variants of the SWiM dataset.

\subsection{Processing Existing Datasets for Spacecraft Segmentation}  

\subsubsection{\textit{PoseBowl}}
\textit{PoseBowl} was created primarily using NASA's Toolset for Training and Labeling in an Optical Simulator (TTALOS). The main workflow of TTALOS leverages high-fidelity 3D spacecraft models to produce annotated, photorealistic training data for computer vision applications. These 3D spacecraft models originate primarily from NASA, with a significant subset contributed by European Space Agency (ESA) for broader coverage of space missions, including spacecraft models such as Rosetta, Aqua, Aura, Mars Odyssey, Cassini, and Voyager.

\textit{PoseBowl} consists of images with only bounding boxes of spacecraft without segmentation masks. These bounding boxes are provided in YOLO format, i.e., ($x_{\text{center}}$, $y_{\text{center}}$, width, height). To address inconsistencies in image formats, we converted all images to JPG format. Then, we generate spacecraft segmentation masks for each image in \textit{PoseBowl} using the Segment Anything Model 2 (SAM 2)~\cite{ravi2024sam} as shown in Figure~\ref{fig:posebowl_example}. SAM 2 provides high-quality segmentation masks with minimal user input, allowing more efficient mask generation in complex images compared to earlier models, such as SAM \cite{ravi2024sam}. Its flexibility and ability to generalize across diverse objects make it more versatile than other models tailored for specific tasks. Since SAM 2 requires bounding box coordinates in the PASCAL Visual Object Classes (VOC) format, i.e., $(x_{\text{min}}, y_{\text{min}}, x_{\text{max}}, y_{\text{max}})$ coordinates, we convert the bounding boxes to the VOC format before applying SAM 2 to obtain binary masks. Finally, these binary masks were converted to polygon coordinates of contours in accordance with the YOLO format.

\subsubsection{\textit{Spacecrafts}}
\begin{figure}[t]
    \centering
    \includegraphics[width=\columnwidth]{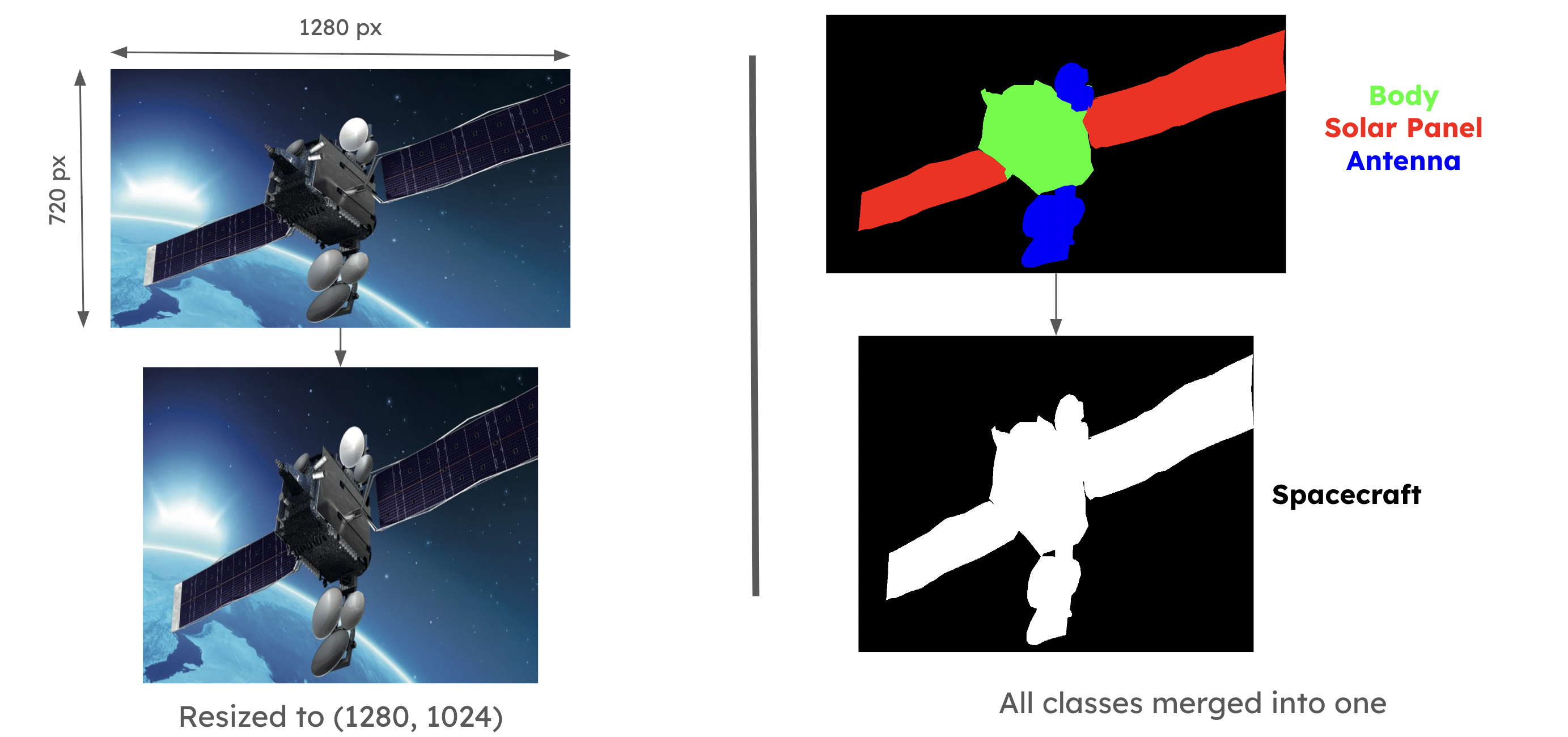}  
    \caption{Example showing resizing an image and merging three segmentation masks into one mask for the \textit{Spacecrafts} dataset.}
    \label{fig:spacecraft_example}
\end{figure}

In the \textit{Spacecrafts} dataset, each image provides a clear and detailed view of the entire spacecraft, accompanied by segmentation masks for three distinct spacecraft regions -- antenna, body, and solar panel. To ensure consistency, we resized all spacecraft images to 1280 $\times$ 1024, matching the image dimensions in \textit{PoseBowl}. Additionally, all three spacecraft region masks were merged into one binary segmentation mask for standardization, as shown in Figure~\ref{fig:spacecraft_example}. 

Finally, the binary masks were reformatted to align with the YOLO format that uses mask contours in the form of polygon coordinates, ensuring compatibility across the SWiM dataset. Since the dataset does not closely resemble real-world data in several cases, we only use these images in the training partition of SWiM to potentially enhance model robustness and generalizability. 

\subsection{Generating Synthetic Images for Our SWiM Dataset}
To create a larger and more comprehensive dataset that can be used to train models for spacecraft segmentation, we generated synthetic images containing spacecraft in addition to preprocessed images from \textit{PoseBowl} and \textit{Spacecrafts} datasets. In particular, we focused on \textit{PoseBowl} \textit{Spacecrafts} images to create synthetic images encompassing more diverse backgrounds and incorporating real-world camera distortions and noise. Our goal in generating these images was to enhance the generalizability and robustness of segmentation models across diverse backgrounds by exposing the models to a wide range of scenarios during training.

\subsubsection{Synthetic Image Generation Methods}

Our synthetic dataset generation included three primary aspects:

\begin{itemize}
    \item Scene Construction and Backgrounds: The pipeline uses real satellite-captured images as backgrounds for scene rendering, creating high realism in composite images. However, synthetically generated backgrounds are also employed to introduce noise and randomness, further enhancing the diversity and robustness of the dataset. This enables superimposition of realistic 3D spacecraft over both authentic and artificial 2D backgrounds, creating variance desirable for training robust machine learning models.
    \item Spacecraft Positioning, Orientation, and Lighting:  The rendering script dynamically positions and rotates the 3D spacecraft models within each scene. This is achieved through random transformations, ensuring coverage of a wide range of visual perspectives. The lighting system is a key feature that uses two modes: in one mode, lighting is synchronized with parameters extracted from the real background image metadata (e.g., sun position and intensity); in the other mode, lighting energy and position are randomized, simulating diverse space illumination conditions. This results in images with spacecraft illuminated and shadowed in a number of realistic combinations.
    \item Augmentation and Compositing Effects: The pipeline supports several image augmentations for added realism and domain generalization:
    \begin{enumerate}
        \item Glare: Different types such as fog glow, streaks, simple star, and ghost glare simulate lens artifacts and environmental glare.
        \item Blur: Adjustable via blending nodes, mimicking motion blur or camera focus artifacts.
        \item Exposure: Randomized exposure settings create effects ranging from underexposure to bright highlight clipping.
    \end{enumerate}
    These compositing augmentations are controlled by a  function which randomizes parameters for each rendered image. Camera and environmental effects can be toggled and tuned according to the desired variance in the generated dataset.

    \item Annotation and Data Export: For every rendered image, the pipeline saves both the image file and corresponding bounding box coordinates that locate the spacecraft within the image.
\end{itemize}

We developed two complementary approaches for image generation, both following similar principles of superimposing spacecraft images onto different backgrounds and applying various augmentation techniques. These augmented images are only part of the augmented version of the SWiM dataset.

\subsubsection{Method 1 (Superimposing segmentations with backgrounds for \textit{Spacecrafts} dataset)}

For our first synthetic image generation approach, we utilized existing resources by combining handpicked background images from two public datasets with open source licenses -- ESA/Hubble images ~\cite{esa_hubble} and ESA Sentinel-2 Satellite images~\cite{esa_sentinel} -- with spacecraft images from the training partition of the \textit{Spacecrafts} dataset (2,145 images). The process involved superimposing segmented spacecraft images onto background images after different types of augmentation:
\begin{itemize}
    \item Background Augmentation: 180-degree rotation, horizontal and vertical flipping, contrast adjustment, saturation adjustment, channel shifting, noise addition, and Gaussian blur. We generated a total of $3,422$ background images.
    \item Spacecraft Augmentation: scaling the spacecraft image between 0.1 and 1.0 times its original size, rotating it by a random angle between 0 and 360 degrees, and flipping it horizontally, vertically, or both.
\end{itemize}

In total, using this method, we generated $21,000$ synthetic images for the training partition of our SWiM dataset. No synthetic images were generated using this method for the validation and test partitions as these images do not resemble realistic images in several cases.  

\subsubsection{Method 2 (TTALOS Pipeline with Stable Diffusion for \textit{PoseBowl} dataset)}

In our second approach, we utilized Stable Diffusion, a deep learning model designed for generating high-quality, photorealistic images from text prompts, to create diverse backgrounds for data augmentation \cite{rombach2022highresolutionimagesynthesislatent}. Guided by the eight classes of backgrounds identified in the \textit{PoseBowl} dataset, we employed corresponding prompts to produce relevant backgrounds. A total of 2,450 unique backgrounds were generated, representing all eight classes. These backgrounds were randomly divided into training, validation, and test partitions of SWiM in proportions of 80\%, 10\%, and 10\%, respectively.

For generating foreground spacecraft images, we used NASA's Toolset for Training and Labeling in an Optical Simulator (TTALOS) pipeline and Blender to render 2D projections of unique 3D spacecraft models \cite{blender}. . In addition to NASA spacecraft models, our dataset incorporated a wider selection of spacecraft from multiple agencies and organizations, including the European Space Agency (ESA), Satellite Communication Service (SES) from Luxembourg, and Johns Hopkins University Applied Physics Laboratory (APL). In particular, we leveraged $15$ unique 3D spacecraft renderings in the training, $5$ in the validation, and $5$ in the test partitions of SWiM. Each rendering featured distinct designs and colors. The 3D spacecraft renderings were generated with randomized poses, lighting conditions, and scales, initially placed on transparent backgrounds. Each spacecraft image was then superimposed onto a unique background generated using Stable Diffusion, resulting in composite images with associated target segmentation masks.

To ensure unbiased and reliable evaluation of model generalization, the spacecraft models included in the training, validation, and test partitions of the SWiM dataset are mostly non-overlapping. This design largely prevents the models from overfitting by memorizing specific spacecraft features seen during training, instead encouraging generalization to novel spacecraft designs, colors, and structures. However, a few spacecraft models are deliberately included in both training and validation/test sets as controlled inclusions, which allow for domain adaptation analysis or robustness checks across different lighting and background conditions.

The spacecraft models for each partition are as follows:
\begin{enumerate}
    \item Training Set (15 models): Rosetta, Aqua, Astra, Aura, Deep Space 1, Mars Odyssey, Messenger, TDRS, Ulysses, Pioneer, Juno, MGS, Galileo, Near, and Ibex.
    \item Validation (5 models): Grace, NPP, EPOXI, Dawn, Acrimsat
    \item Test (5 models): Voyager, Themis, Cassini 66, Stardust, MRO
\end{enumerate}
This structured partitioning strategy enables a robust training process while facilitating realistic evaluation scenarios that test the model's ability to perform well on unseen spacecraft configurations.

To introduce visual diversity, we applied further transformations to these composite images, including random changes in poses, lighting, and sizes, as well as geometric adjustments such as rotations, flips, and scaling. Additional visual effects, such as color adjustments and Gaussian blur, were applied to these images to simulate realistic visual distortions encountered in image acquisition in space.

Beyond reusing satellite images, we greatly increase scene complexity by generating custom backgrounds using the diffusion model. This approach inserts rare, unexpected, or extreme conditions that may be underrepresented or absent in satellite-captured images:
\begin{itemize}
    \item Carefully crafted text prompts are used to instruct the diffusion model to create highly varied, high-resolution space backgrounds. These range from faint starlit voids, dense star fields, blurred planetary fly-by, dramatic sun glares, atmospheric auroras, to night-time city lights.
    \item Sensor and atmospheric artifacts, such as chromatic aberration, lens flare, pixelation, and simulated sensor noise, are described in the prompt and infused into the output, resulting in backgrounds that challenge and generalize the training data far beyond ordinary real satellite imagery.
\end{itemize}

In total, we generated $10,000$ synthetic images using this method for the training partition. In addition, we generated $2,000$ synthetic images for each of the validation and test partitions.

\subsection{Capturing Real Imaging Complexity}

Overall, our pipeline is specifically designed to fully capture the unpredictability and complexity of real orbital imaging conditions by integrating both natural and artificially generated variations at every stage of data creation. We capture real imaging complexity through the following means:

\begin {enumerate}
    \item Lighting Variability: We randomize both the intensity and spatial position of the primary light source, mimicking rapid changes in solar illumination and shadowing experienced in orbit.
    \item Camera Effects: Augmentations such as random blur (to simulate motion or out-of-focus artifacts), exposure variation (for under/overexposed scenes), and glare (from sun angles, sensor mechanics, or rapid light transitions) are systematically applied. These imitate the unpredictable optical effects captured by real orbital cameras.
    Diverse Backgrounds: By default, scenes use real satellite images; we further add complexity to this by adding synthetic backgrounds.
\end{enumerate}

\subsection{Diversity of Field of View and Occlusion}

\textit{PoseBowl} features images in which spacecraft typically occupy a small portion of the overall image area; over 85\% of images contain spacecraft that cover less than 10\% of the total image area. In comparison, the Spacecraft dataset is characterized by a more uniform distribution of spacecraft size within images; spacecraft spanning a broader range, from 10\% up to 80\% of the image area. 

Importantly, the baseline Posebowl dataset is approximately eight times larger than the Spacecraft dataset, ensuring that, even after merging, the combined dataset will still contain a substantial number of images with small field-of-view (FoV) spacecraft. This large sample size closely mirrors real-world orbital imaging, where spacecraft are often distant or only partially captured within each frame, preserving a crucial aspect of operational realism. Nonetheless, the inclusion of a broader range of spacecraft sizes from the Spacecraft dataset is highly beneficial -- it introduces valuable variability that supports robust model generalization and improves segmentation performance across both small and large FoVs, ultimately aligning the data distribution more closely with the full spectrum of scenarios encountered in practice.

Moreover, the presence of occlusion in our dataset allows segmentation models to learn how to identify and delineate spacecraft even when they are partially blocked by other objects or image artifacts, closely reflecting real-world orbital imaging scenarios. This strengthens the model’s ability to recognize the full extent of objects within cluttered or challenging visual environments,

\subsection{SWiM Dataset Versions}

The SWiM data is available in two versions, baseline and augmented. Both versions have separate training, validation, and test splits. The details of the two versions are provided here.

\subsubsection{Baseline}

The training partition of the baseline dataset consists of $18,399$ images and annotation corresponding to the training set of \textit{PoseBowl} and the entirety of the \textit{Spacecrafts} dataset for a total of $21,515$ images with segmentation masks. The validation and test partitions each consist of $3,701$ images and annotation corresponding to the validation and test images in the \textit{PoseBowl} dataset. The complete dataset contains $28,917$ images with corresponding segmentation masks.

\subsubsection{Augmented}

\begin{figure}[t]  
    \includegraphics[width=\columnwidth]{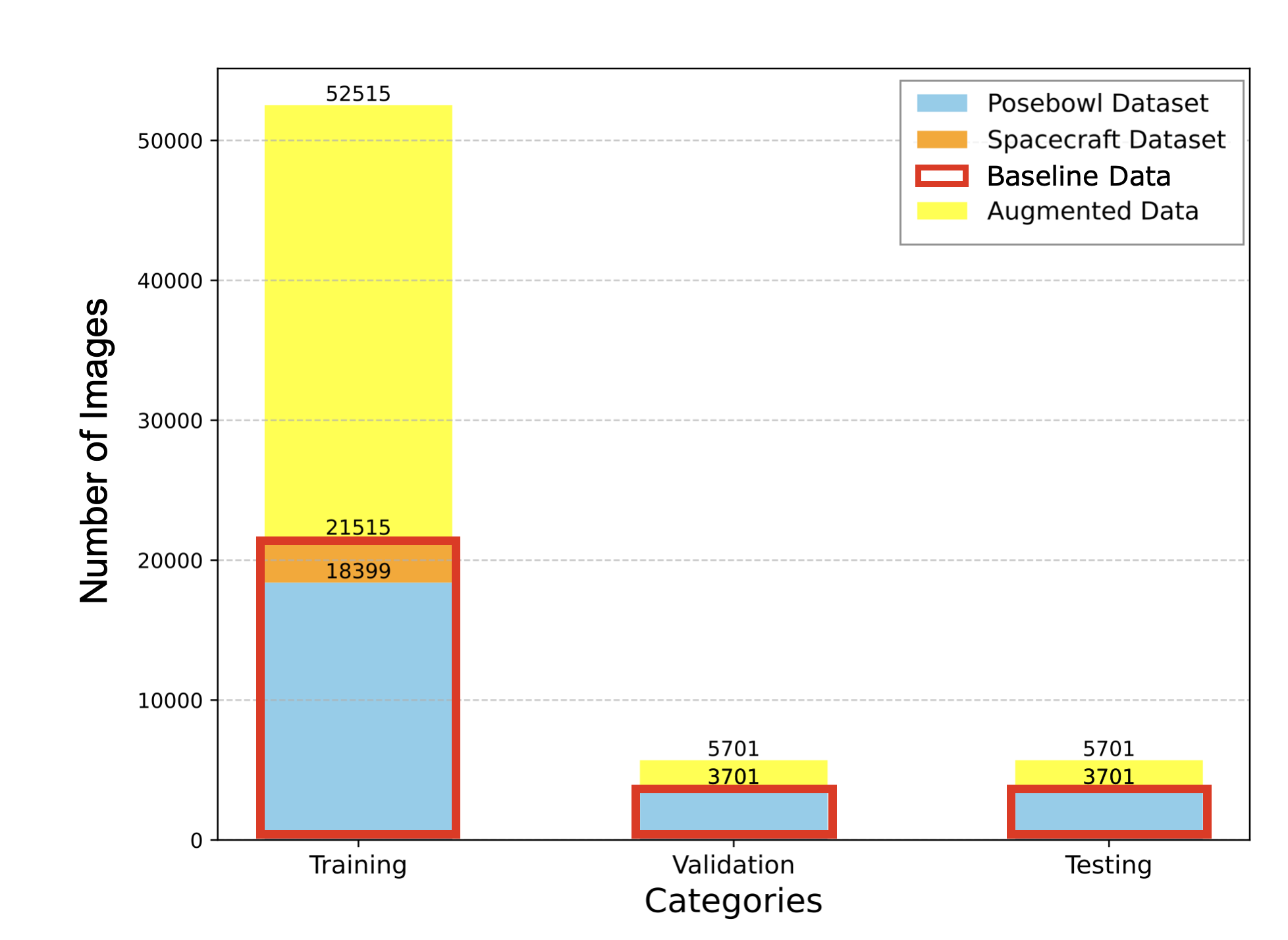}
    \caption{Number of images in the baseline and augmented versions of SWiM.}
    \label{data_split}
\end{figure}

\begin{figure*}[t]  
    \begin{subfigure}{0.19\textwidth}
        \centering
        \includegraphics[width=.9\linewidth]{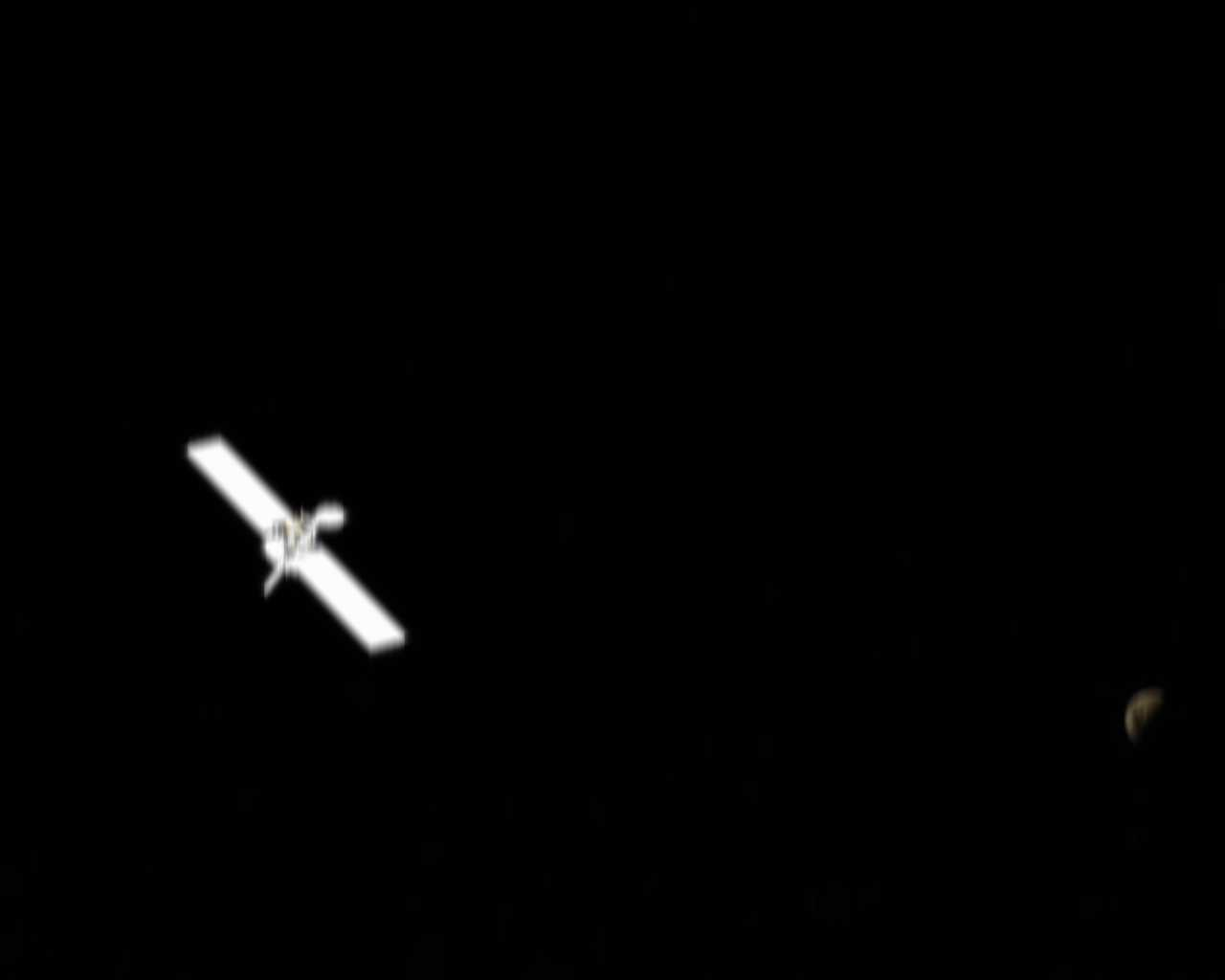}
    \end{subfigure}%
    \begin{subfigure}{0.19\textwidth}
    \centering
        \includegraphics[width=.9\linewidth]{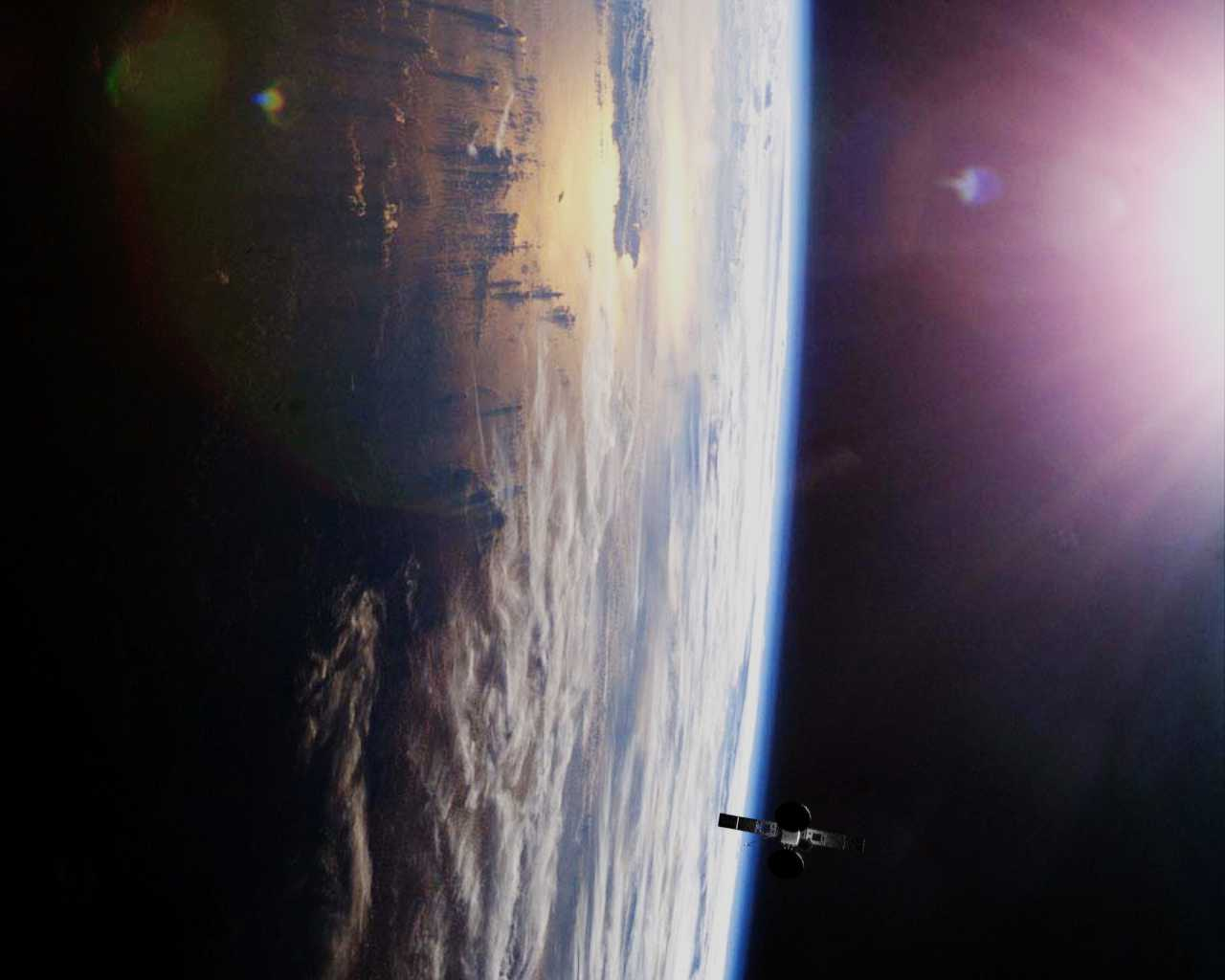}
    \end{subfigure}
    \begin{subfigure}{0.19\textwidth}
    \centering
        \includegraphics[width=.9\linewidth]{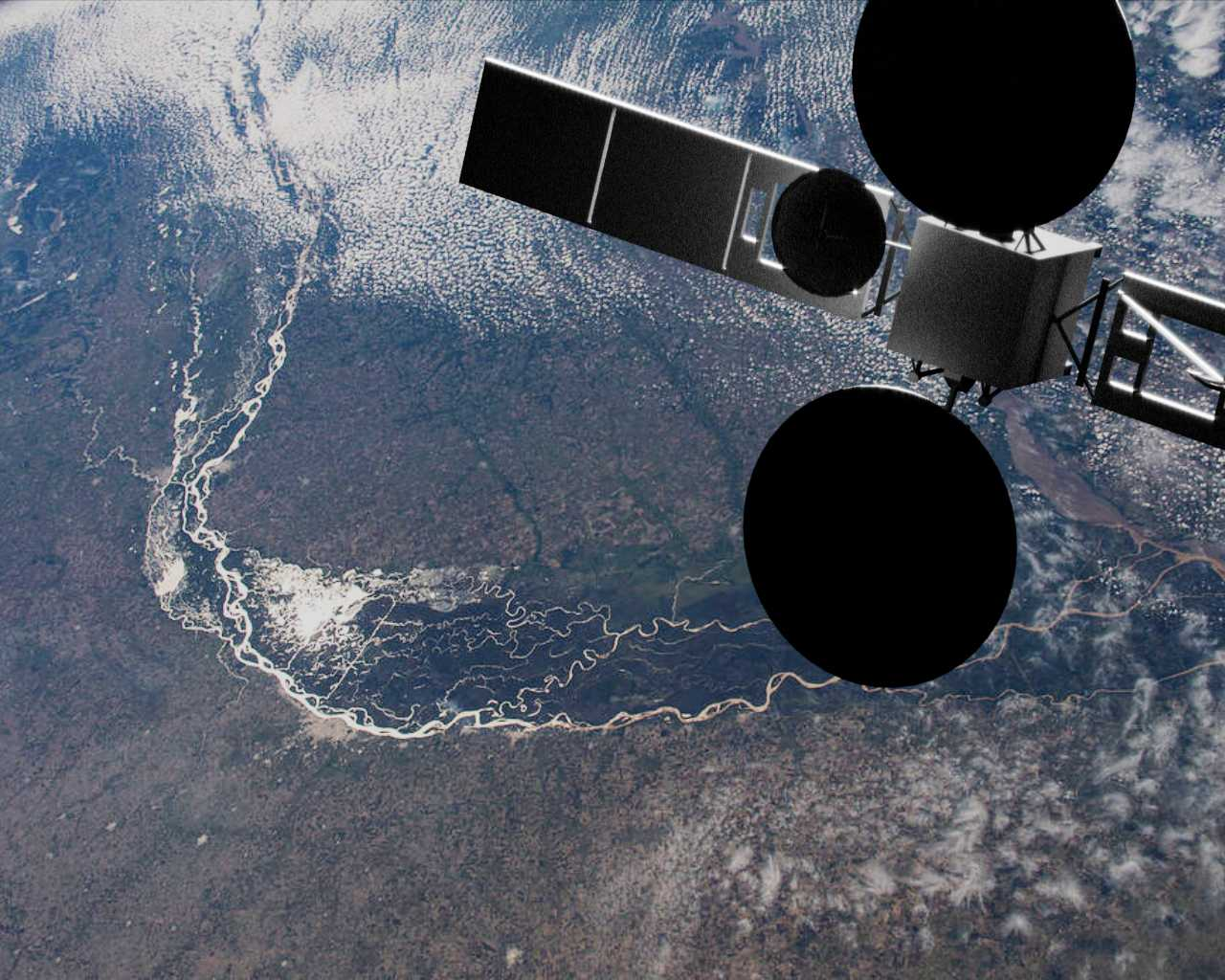}
    \end{subfigure}
    \begin{subfigure}{0.19\textwidth}
    \centering
        \includegraphics[width=.9\linewidth]{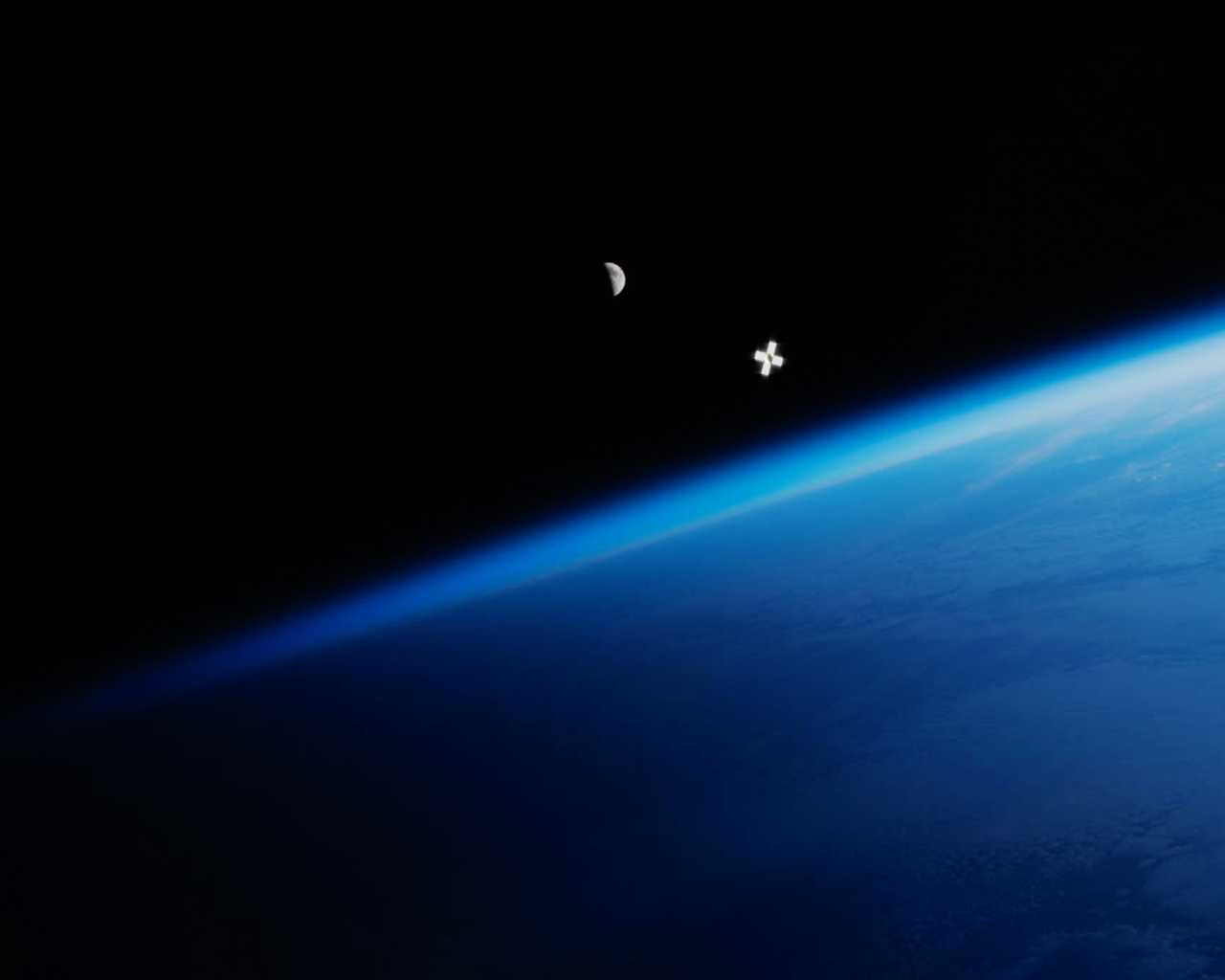}
    \end{subfigure}
    \begin{subfigure}{0.19\textwidth}
    \centering
        \includegraphics[width=.9\linewidth]{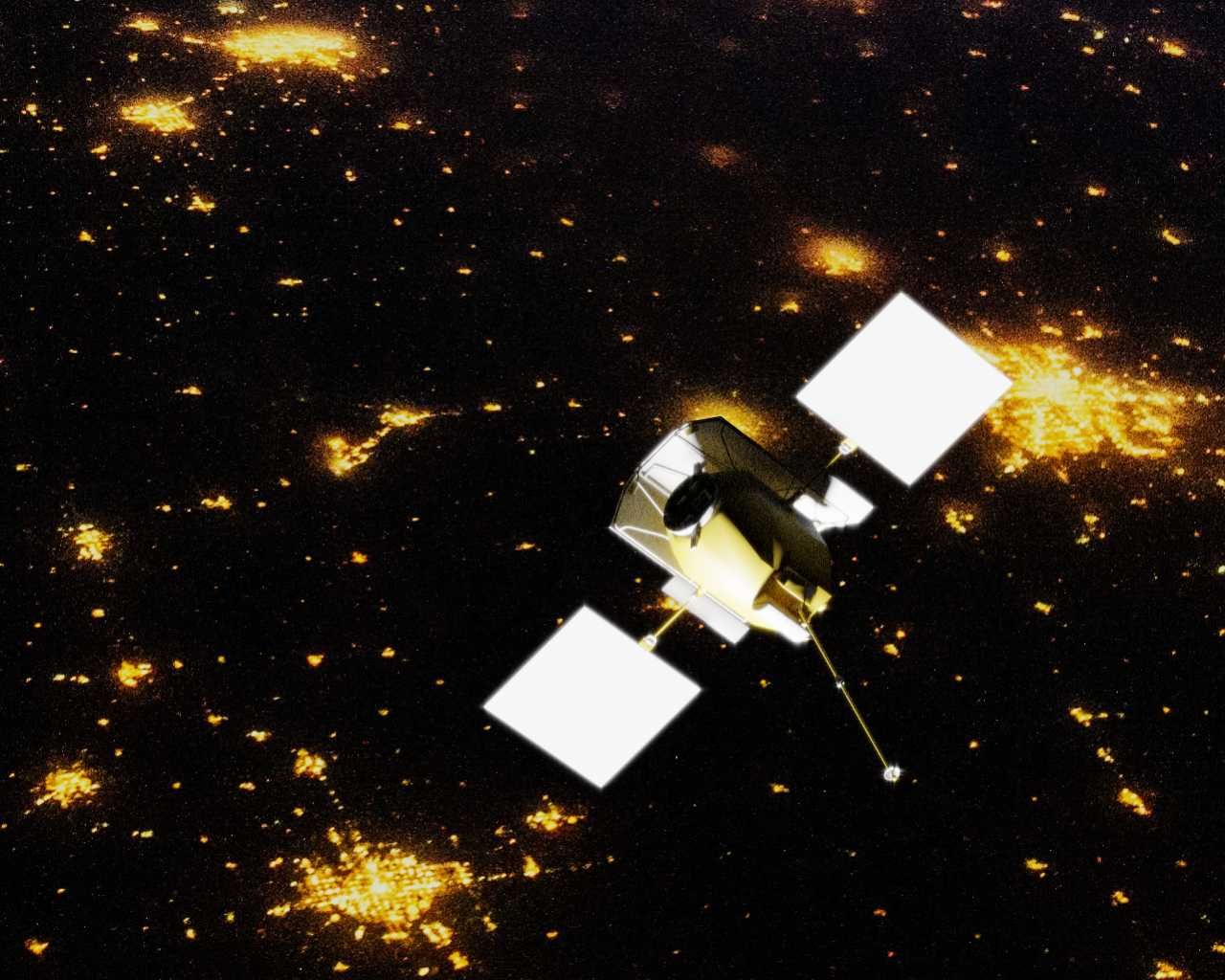}
    \end{subfigure}
    \caption{Sample synthetic images in the augmented dataset. Note the diversity in background and the real-world challenges, such as glare and confounding background.}
    \label{sample_augmented}
\end{figure*}

The training partition of the augmented dataset comprises the training partition of the baseline dataset along with the $21,000$ synthetically generated images using Method 1 and $10,000$ images generated using Method 2. In total, the training set contains $52,515$ images. The validation and test partitions consist of the validation and test partitions in the baseline version appended with additional $2,000$ images generated using Method 2 for each partition. Thus, both the validation and test partitions consist of $5,701$ images. The complete dataset consists of $63,917$ images with segmentation masks.

In this work, we define ``synthetically generated images'' as composite samples created by generating virtual backgrounds via Stable Diffusion and integrating them into the TTALOS pipeline to superimpose 3D spacecraft models. These synthetic composites are exclusively allocated to the training partition to enhance environmental diversity. Conversely, the validation and test partitions consist entirely of images featuring real orbital backgrounds derived from the original \textit{PoseBowl} dataset. This stratification ensures that while the model is exposed to high-variance synthetic domains during training, it is benchmarked strictly against photorealistic imagery representative of actual operational conditions.

Figure~\ref{data_split} shows the detailed composition and sizes of the training, validation, and test partitions of both the baseline and augmented versions of SWiM. Some examples of the synthetic images used in the augmented dataset are shown in Figure~\ref{sample_augmented}. We note the deliberate incorporation of a variety of realistic backgrounds, including blank dark space with and without moons and planets, blurred images of planets simulating motion blur of spacecraft, glare from sunlight, and earth's terrain during day and night. 

\section{Models for Performance Benchmarks}

Considering the strict constraints on hardware and inference time, we selected the YOLO family of models over other models, such as Mask-RCNN~\cite{maskrcnn} and transformer-based models~\cite{transformer_seg}, for generating benchmark performance results using the SWiM dataset. In particular, we chose YOLOv8~\cite{ultralytics_yolov8} and YOLOv11~\cite{ultralytics_yolo11}, two of the recent versions in the YOLO family in our experiments to satisfy the constraints. The YOLO family of models, particularly the nano variants, provides an optimal trade-off between model size, inference speed, and segmentation accuracy.

While each YOLO model is available in multiple variants, depending on size and inference time, we chose the nano versions of the models due to the computational and operational constraints imposed by a resource-constrained target deployment environment. The hardware, being a compact single-board computer, offers limited hardware capabilities, including the absence of a GPU, only 4 GB of RAM memory, and 4 CPU cores \cite{upboard_specifications}. Consequently, the segmentation algorithm must not only achieve high accuracy but also perform inference in real-time, ideally under 0.95s per image, to support spacecraft inspection during proximity operations in space \cite{drivendata2024pose}.

YOLOv8 Nano, with approximately 3.4M parameters and a 17 MB RAM footprint at full precision during inference, has already demonstrated successful deployment in space operations. It was validated during NASA’s Pose Bowl Spacecraft Detection challenge, where it achieved real-time object detection on the \textit{PoseBowl} dataset and was subsequently deployed on an operational NASA spacecraft \cite{drivendata2024pose}. YOLOv8 Nano’s architecture is highly compatible with resource-constrained environments, making it a top candidate to extend from object detection to instance segmentation tasks. The architectural improvements of YOLOv11 also present a promising alternative with even better inference times.

\begin{figure*}[t]
    \centering
    \includegraphics[scale=0.3]{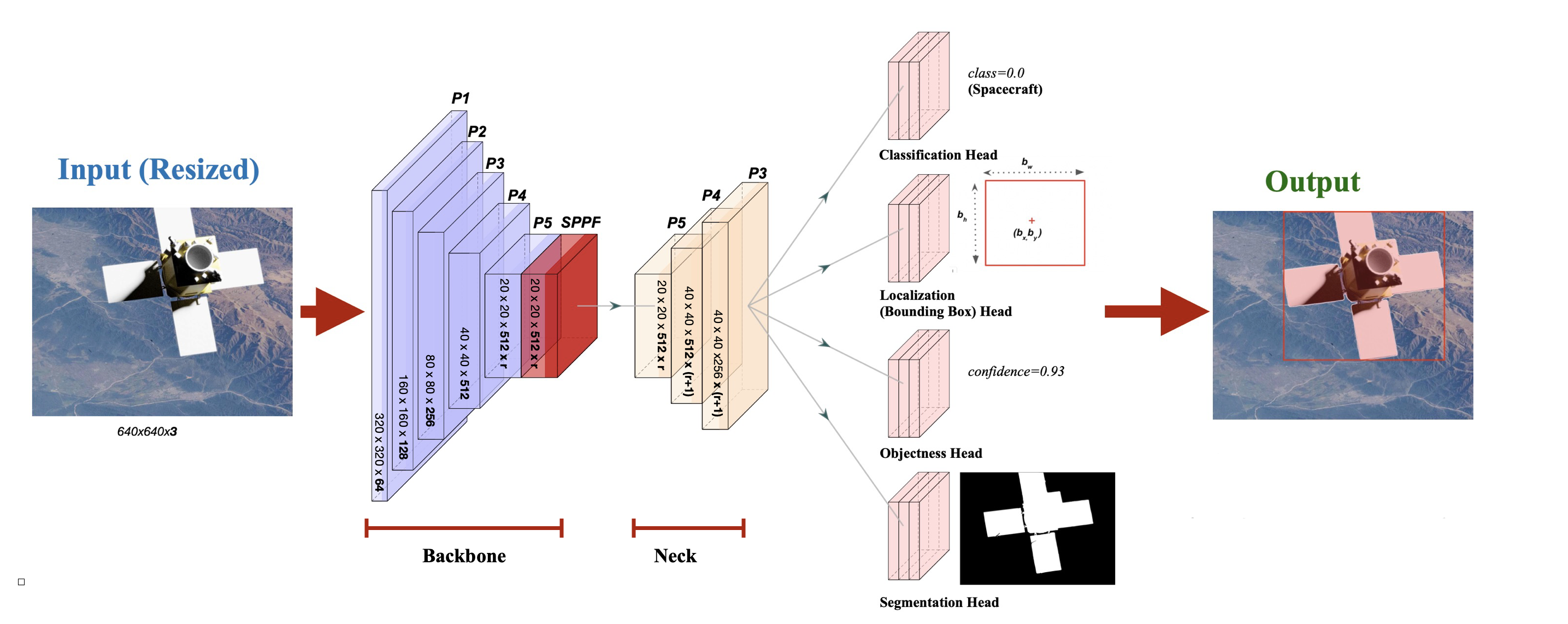}
    \caption{Overview of the YOLO Nano model architecture. The diagram illustrates the model's key components, including the backbone, the neck, and the four decoupled heads (classification, localization, objectness, and segmentation). The input image (resized to 640×640×3) passes through the backbone and neck, where features are processed and routed to the respective heads.}
    \label{block_arc}
\end{figure*}

Further, the nano variant's lower power consumption is a critical factor in space-based applications where energy efficiency is paramount. Most importantly, YOLOv8's small variant, which is slightly larger (7M parameters) than the nano variant had an inference time greater than 1s and thus does not meet the design constraints. Thus, the adoption of YOLOv8 Nano and YOLOv11 Nano as benchmarks for segmentation performance for our \textit{SWiM} dataset ensures alignment with NASA’s hardware limitations and real-time performance requirements.

The YOLOv8-seg and YOLOv11-seg nano models~\cite{ultralytics_yolov8, ultralytics_yolo11} operate by predicting bounding boxes and associated masks for segmentation. The YOLO model architecture is organized into three primary components: backbone, neck, and head, each performing distinct functions to achieve efficient object detection and segmentation through four decoupled output heads. In our experiments, we only leveraged the segmentation head of the model. YOLO Nano, a lightweight variant, is designed for speed and efficiency, with reduced depth and width to accommodate resource-constrained environments. Figure \ref{block_arc} highlights the key components in the YOLO nano segmentation architecture.

\section{Experiments and Results}
\subsection{Evaluation Metrics}\label{AA}
Benchmark performance is reported using the following metrics:
\subsubsection{Dice Coefficient:} The Dice coefficient is a widely used metric in image segmentation that quantifies the overlap between the predicted segmentation mask and the ground truth mask. It is calculated as twice the area of overlap divided by the total number of pixels in both masks.
    The formula for the Dice coefficient is:
    \begin{equation}
        \text{Dice} = \frac{2 * |X \cap Y|}{|X| + |Y|}
    \end{equation}
    where $X$ is the predicted set of pixels, $Y$ is the ground truth set of pixels $|X \cap Y|$ represents the area of overlap between $X$ and $Y$, and $|X| + |Y|$ are the total number of pixels in $X$ and $Y$.
    
    Dice coefficient ranges from 0 to 1, with 1 indicating perfect overlap between the predicted and ground truth masks. It emphasizes the importance of true positive predictions, making it particularly useful for assessing segmentation accuracy, and is less sensitive to outliers compared to Intersection over Union (IoU). Moreover, it works well across all segmentation mask sizes, especially for small masks, making it particularly suitable for evaluating spacecraft masks that are relatively small compared to the overall image area.


\subsubsection{Hausdorff Distance :} Hausdorff Distance is a metric used to measure the similarity between two sets of points, particularly useful in image segmentation for comparing the boundaries of predicted and ground truth masks. It quantifies the maximum distance between the corresponding points at the boundaries of ground truth and predicted segmentation masks. 
    
    The directed/asymmetric Hausdorff distance from set X to set Y is defined as:
    \begin{equation}
    h(X,Y) = \max_{x \in X} \{\min_{y \in Y} d(x,y)\}
    \end{equation}


    In our experiments, we use the symmetric Hausdorff Distance, defined as:

    \begin{equation}
    H(X,Y) = \max\{h(X,Y), h(Y,X)\}
    \end{equation}
    
    where $X$ is the predicted set of pixels, $Y$ is the ground truth set of pixels, and $d(x,y)$ is the Euclidean distance between points $x$ and $y$. We note here that Hausdorff Distance provides insight into the worst-case segmentation error and is particularly useful for evaluating the accuracy of object boundaries in segmentation tasks. In particular, Hausdorff Distance evaluates the boundaries of the segmentation masks and is not a simple region-based calculation that is agnostic to the boundaries of masks, such as Dice Coefficient. Thus, it complements the Dice coefficient by focusing on the spatial boundaries of segmentation masks rather than a simple volumetric overlap.

\subsection{Experiments}
To ensure that the YOLOv8 and YOLOv11 models could be deployed in the resource-constrained environment of a spacecraft, we optimized their resource requirements using quantization techniques through ONNX Runtime~\cite{onnxruntime} and exported the models to ONNX format due to its proven performance in NASA's Pose Bowl competition~\cite{drivendata2024pose}.

ONNX (Open Neural Network Exchange) is a widely adopted format that enables the interoperability and portability of machine learning models in different frameworks and hardware platforms~\cite{onnxruntime}. This compatibility is particularly beneficial for scenarios that require deployment across diverse environments, such as edge devices or resource-constrained systems. ONNX Runtime complements the ONNX format through a high-performance inference engine optimized for various hardware configurations~\cite{graph}, resulting in key advantages, such as cross-platform compatibility, performance optimization (e.g., graph optimizations, operator fusions, and kernel tuning), resource efficiency, and quantization support~\cite{onnxruntime}.

Further, to simulate a typical resource-constrained onboard flight computer with 4GB RAM, a docker image and container were created, and limited to 3 GB RAM and 3 CPU cores, leaving aside the remaining hardware for contingency. This extreme buffer design choice ensured that the model was tested under tighter constraints for inference time.

\subsection{Results}
\subsubsection{Performance Benchmark for Segmentation}
\begin{table}[t]
    \caption{Performance Benchmark using YOLO v8n and YOLO v11n trained on baseline and augmented versions of SWiM. (Base stands for Baseline; Aug stands for Augmented)}
    \centering
    \begin{tabular}{|c|c|c|c|}
    \hline
    \textbf{Model} & \begin{tabular}{@{}c@{}}\textbf{SWiM} \\ \textbf{Version}\end{tabular} & \begin{tabular}{@{}c@{}}\textbf{Average} \\ \textbf{Dice Score}\end{tabular} & \begin{tabular}{@{}c@{}}\textbf{Average} \\ \textbf{Hausdorff}\end{tabular}\\
    \hline
    \hline
    YOLO v8n & Base & 0.9292 & 0.6895\\
    \hline
    YOLO v8n & Aug & 0.9269 & 0.8857\\
    \hline
    YOLO v11n & Base & 0.9287 & 1.0781\\
    \hline
    YOLO v11n & Aug & 0.9274 & 0.7476\\
    \hline
    \end{tabular}
    
    \label{metric table}
\end{table}

Table \ref{metric table} presents the test performance of the two versions of the YOLO segmentation model, YOLOv8 nano and YOLOv11 nano trained on the baseline and augmented training sets using Dice coefficient and Hausdorff Distance. We note that all the models achieved a high average Dice score of $0.92$ on the test dataset, indicating that the masks are generally very accurate in identifying the bulk area of the spacecraft. All 4 models have a very low average Hausdorff distance between $0.69$-$1.07$, indicating that the models are generally excellent at detecting the complex boundaries of various spacecraft models. 

Overall, YOLO v8 and v11 both achieve excellent performance in detecting the bulk of the spacecraft, as well as the boundaries of most spacecraft. We note that using the augmented version of SWiM does not necessarily have a significant impact on the performance of YOLO. 

The consistent accuracy for both YOLO v8 and v11 verifies that both the datasets result in highly accurate models across different YOLO architectures. The difference in Dice coefficient between baseline and augmented datasets is of the order of $10^{-4}$, which indicates that the YOLO models are robust to the additional noise added in the augmented dataset. As this noise mimics several real-world scenarios such as camera errors and lighting issues, this robustness is very promising. 

Based on the success of YOLO architectures on both datasets, we propose that the baseline and augmented datasets provide two options in terms of size for future work using other model architectures.

\subsubsection{Inference Time and Model Size}
The inference times of YOLOv8 nano and YOLOv11 nano were recorded in the ONNX format under resource-constrained conditions for benchmarking. The benchmark for inference time and model size (Table \ref{inference-table}) includes inference time per image, frames per second (FPS), and the model size (MB), highlighting the computational efficiency, and storage requirements.  

\begin{table}[ht]
    \caption{Benchmark Inference Time for Models. (Base stands for Baseline; Aug stands for Augmented)}
    \centering
    \begin{tabular}{|c|c|c|c|c|}
    \hline
    \textbf{Model} & \begin{tabular}{@{}c@{}}\textbf{SWiM} \\ \textbf{Version}\end{tabular} & \textbf{Inf. (ms)} & \textbf{FPS} & \textbf{Size (MB)}\\
    \hline
    \hline
    YOLO v8n & Base & 443.90 & 2.25 & 12.7\\
    \hline
    YOLO v8n & Aug & 586.78 & 1.87 & 12.7\\
    \hline
    YOLO v11n & Base & 533.62 & 1.7 & 11.1\\
    \hline
    YOLO v11n & Aug & 446.93 & 2.24 & 11.1\\
    \hline
    \end{tabular}
    \label{inference-table}
\end{table}

The models successfully achieved near real-time inference speeds, with an average inference time of 500 ms, which is nearly half the goal time constraint of 0.95s. Though the YOLO v11 claims a more optimized model architecture, we observed only minor differences in the performances of v8 vs v11. These differences can be attributed to resource allocation processes during the simulation. Thus, both the v8 and v11 prove to be excellent choices based on performance as well as accuracy. 

\section{Discussion}
In this work, we introduce a large dataset for spacecraft segmentation by standardizing two existing datasets in terms of size and ground truth mask annotations as well as creating synthetic but realistic images along with segmentation masks. Our dataset, \textit{SWiM} (Spacecraft With Masks) is available in two versions, baseline ($28,917$ annotated images) and augmented ($63,917$ annotated images). While the baseline version contains images with standardized segmentation masks from the existing datasets, the augmented version additionally consists of synthetic images generated using two different methods, involving the use of images from European Space Agency satellite imagery as well as those generated using stable diffusion. For the purposes of model comparison, our dataset is pre-split into training, validation, and test partitions. To make the performance evaluation realistic, the validation and test partitions for both baseline and augmented versions do not contain any synthetic images.

In addition, we define the constraints for model development considering deployment on resource-constrained onboard hardware in space flight and inference time limits for real-time operation (based on the current operating specifications of NASA). Using these constraints, we obtain and report benchmark performance results using YOLOv8 nano and YOLOv11 nano models. We note that YOLOv11 nano achieves comparable inference speed to YOLOv8 nano while being more storage-efficient, aligning with the project’s objectives for efficient, edge-compatible deployment. Both models successfully met all technical constraints of on-board deployment by NASA, operating efficiently within a constraint-specific Docker container ($<$4GB RAM with 4 cores and CPU-only inference) and achieving an inference time under 0.95s per image, as required for real-time performance in autonomous spacecraft inspection. 

We also introduce an evaluation protocol using two different metrics for performance evaluation -- Dice score and Hausdorff distance -- stressing the need to incorporate both region-based and boundary-aware evaluation of predicted segmentation masks. In conclusion, we hope that the impact of our work, involving a new large annotated dataset, definition of constraints for onboard autonomous inspection in space, performance benchmark, and assessment protocol, will help researchers and scientists develop models for more real-time vision applications in space exploration.

We also note certain limitations of our dataset. For segmentation mask generation, SAM2 was fine-tuned on a small sample of \textit{PoseBowl}, in which the majority of the images had a field of view (FOV) less than 0.1 of the area of the image. Consequently, the SAM2 masks are likely to be the most accurate for similar images. For the 509 images with an FOV of over 0.5, we found that 98 presented inaccuracies that we manually corrected. The inaccurate masks produced by SAM2 primarily exhibited two common themes: misclassification of background pixels and edge noise. In some cases, the mask includes pixels from the background, treating them as part of the spacecraft structure. In others, the edges of the spacecraft are poorly defined, leading to imprecise segmentation boundaries.

The SWiM dataset and the software for performance benchmark are available at \url{https://github.com/RiceD2KLab/SWiM}.

\bibliographystyle{IEEEtran}
\bibliography{iclr2023_conference}

\end{document}